\pdfoutput=1 

\documentclass[journal]{IEEEtran}
\usepackage{cite}

\ifCLASSINFOpdf
\else
\fi
\usepackage{amsmath}
\usepackage{amsfonts}
\usepackage{times}
\usepackage{epsfig}
\usepackage{graphicx}
\usepackage{amsmath}
\usepackage{amssymb}
\usepackage{booktabs}
\usepackage{multirow}
\usepackage[table,xcdraw]{xcolor}
\usepackage{bm} 
\usepackage{makecell}
\usepackage{threeparttable}

\usepackage{algorithmic}

\usepackage{threeparttable}
\usepackage{multirow}
\usepackage{makecell}
\usepackage{algorithm}
\usepackage{algorithmic}
\usepackage{url}
\usepackage{subfigure}
\usepackage{graphicx}
\usepackage{color}

\newcommand{\eg}{\emph{e.g}.}
\newcommand{\etal}{\emph{et al}.}

\newcommand{\PreserveBackslash}[1]{\let\temp=\\#1\let\\=\temp}
\newcolumntype{C}[1]{>{\PreserveBackslash\centering}p{#1}}
\newcolumntype{R}[1]{>{\PreserveBackslash\raggedleft}p{#1}}
\newcolumntype{L}[1]{>{\PreserveBackslash\raggedright}p{#1}}
\newcolumntype{M}[1]{>{\centering\arraybackslash}m{#1}}

\begin{document}
\bstctlcite{IEEEexample:BSTcontrol}
%
\title{Variational Probabilistic Fusion Network for RGB-T Semantic Segmentation}
%
%
%


\author{Baihong Lin*,
		Zengrong Lin*,
		Yulan Guo,
		Yulan Zhang,
		Jianxiao Zou,
		Shicai Fan

\thanks{Manuscript received August 19, 2023; revised August 16, 2023. The work is supported in part by Natural Science Foundation of Guangdong Province (China) under Grant 2022A1515010493, in part by the Fundamental Research Funds for the Central Universities (Sun Yat-sen University) under Grant 2021qntd11, in part by the Science Technology and Innovation Commission of Shenzhen Municipality under Grant 2021A03, in part by the Guangdong Basic and Applied Basic Research Foundation (2022B1515020103).}
\thanks{B. Lin, J. Zou and S. Fan are with the Institute for Advanced Study, University of Electronic Science and Technology of China, Shenzhen, 518107, China. Z. Lin, Y. Guo and Y. Zhang are with the School of Electronics and Communication Engineering, the Shenzhen Campus, Sun Yat-sen University, Shenzhen, 518107, China.}
\thanks{*B. Lin is the first author and corresponding author. \{E-mail: linbaihong111@126.com\}. Z. Lin is the co-first author.}
}

%
%

\markboth{Journal of \LaTeX\ Class Files,~Vol.~14, No.~8, August~2015}%
{Shell \MakeLowercase{\textit{et al.}}: Bare Demo of IEEEtran.cls for IEEE Journals}
%



\maketitle

\begin{abstract}
RGB-T semantic segmentation has been widely adopted to handle hard scenes with poor lighting conditions by fusing different modality features of RGB and thermal images. Existing methods try to find an optimal fusion feature for segmentation, resulting in sensitivity to modality noise, class-imbalance, and modality bias. To overcome the problems, this paper proposes a novel Variational Probabilistic Fusion Network (VPFNet), which regards fusion features as random variables and obtains robust segmentation by averaging segmentation results under multiple samples of fusion features. The random samples generation of fusion features in VPFNet is realized by a novel Variational Feature Fusion Module (VFFM) designed based on variation attention. To further avoid class-imbalance and modality bias, we employ the weighted cross-entropy loss and introduce prior information of illumination and category to control the proposed VFFM. Experimental results on MFNet and PST900 datasets demonstrate that the proposed VPFNet can achieve state-of-the-art segmentation performance.
\end{abstract}

\begin{IEEEkeywords}
RGB-T semantic segmentation, probabilistic fusion, modality noise, class-imbalance, modality bias
\end{IEEEkeywords}

%
\IEEEpeerreviewmaketitle

\section{Introduction}
\label{sec:intro}
RGB-based semantic segmentation aims at categorizing each pixel in a RGB image into a specific class, which has been applied in many domains, including robot vision, autonomous driving, and path planning~\cite{drive2,drive1,robot1,robot2}. Unfortunately, a RGB image is sensitive to illumination, thus, recent studies propose RGB-T semantic segmentation, i.e., semantic segmentation based on the fusion of paired RGB and thermal images, to deal with poor illumination caused by low-light, rainy, and foggy conditions~\cite{MFNet,EGFNet,ABMDRNet}.

So far, deep learning methods have been widely adopted for RGB-T semantic segmentation, since they usually work excellently in many existing datasets. To achieve good segmentation performance, it is important that how to effectively fuse the deep complementary features from both RGB and thermal images. Early studies directly fuse deep features of RGB and thermal images by simply adopting element-wised addition or concatenation. But these designs ignore the modality differences between RGB and thermal images, leading to inadequate cross-modality complementary information exploitation~\cite{ABMDRNet}. To solve the above problem, recent studies provide different fusion designs: Some researchers introduce attention mechanisms for features of RGB and thermal images before fusion to reduce the modality differences~\cite{FEANet,doodlenet,ABMDRNet}; others leverage prior information, such as object boundaries and contours, to guide the feature generation and fusion processes, thereby preserving relevant information from both modalities~\cite{EGFNet,zhou2021gmnet,MFFENet}. These fusion designs have been demonstrated to enhance the performance of RGB-T semantic segmentation.

\begin{figure}
	\centering
	\includegraphics[width=83mm]{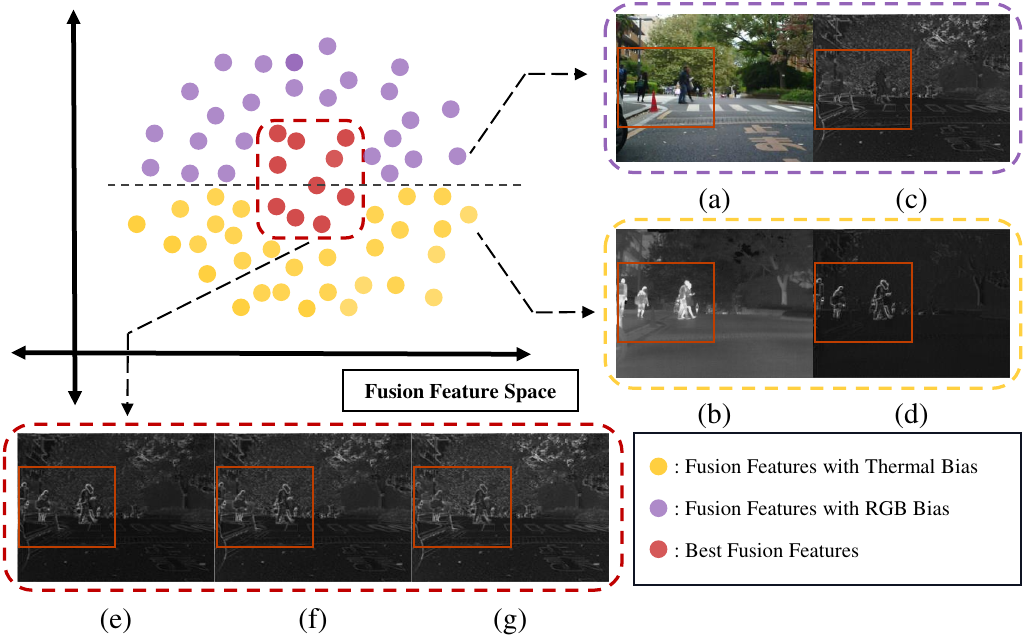}
	\caption{Illustration of the assumption that the optimal features for semantic segmentation are a series of fusion features within a certain range. Circles represent fusion features. (a)-(d) Original RGB image, thermal image and their features, respectively. (e)-(g) A series of different fusion features which can achieve equally good semantic segmentation performance. }
	\label{fig:sample}
\end{figure}


However, existing methods are limited by the implicit assumption that the optimal fusion feature of RGB and thermal images is unique for the following reason: First, the unique fusion assumption will weaken generalization ability, since the fusion feature is usually vulnerable to class-imbalance~\cite{imbalancereview}, modality bias~\cite{peng2022balanced}, and modality noise~\cite{chen2020bi}. Second, this assumption limits feature representation ability, thus, most fusion designs based on this assumption work badly without RGB or thermal images.


In this paper, instead of continuing previous works to find an optimal fusion feature for segmentation, we propose the \emph{Variational Probabilistic Fusion Network} (VPFNet) for RGB-T semantic segmentation, which assumes that the optimal features for semantic segmentation are a series of fusion features within a certain range as shown in Fig.\ref{fig:sample} to alleviate the influence of modality noise, class-imbalance, and modality bias. 
Specifically, inspired by \cite{VAE,chen2021variational}, we regard fusion features as random variables and propose a novel fusion module, i.e., \emph{Variational Feature Fusion Module} (VFFM), to generate random samples of fusion features for segmentation. 
To avoid class-imbalance and modality bias, we employ the weighted cross-entropy loss and introduce Gaussian Mixture distribution as the prior of category and illumination for the proposed VFFM under variation framework.
Using the above modeling, the final segmentation mask is predicted by averaging all segmentation confidence maps calculated from different random samples of fusion features to reduce sensitivity to modality noise.
In brief, the contributions of this paper are listed as follows:
\begin{itemize} 
	\item {} VPFNet is proposed to obtain robust segmentation by regarding fusion features as random variables and averaging segmentation results under multiple samples of fusion features, which can alleviate the influence of modality noise, class-imbalance, and modality bias. 
	\item {} VFFM is designed based on variation attention, and is controlled by category and illumination priors to generate random samples of fusion feature for multiple segmentation results. To our knowledge, it is the first work to break the unique fusion assumption.
	\item {} Experiments on MFNet and PST900 datasets demonstrate that the proposed fusion network achieves excellent segmentation performance compared with other state-of-the-art methods.
\end{itemize}

\section{Related Work}
\label{sec:relatedwork}

\subsection{RGB-based Semantic Segmentation}

Traditional RGB-based semantic segmentation methods utilized random forest~\cite{breiman2001random} and conditional random field~\cite{lafferty2001conditional} to construct pixel-level classifiers to realize semantic segmentation based on hand-crafted features, such as histograms of oriented gradients~\cite{dalal2005histograms} and scale-invariant feature transform~\cite{lindeberg2012scale}.

Recently, deep learning methods have become the dominant solutions for RGB-based semantic segmentation since these methods can easily achieve remarkable results based on machine learnt features. Long \etal~\cite{FCN} first propose an efficient end-to-end Fully Convolutional Network for semantic segmentation to replace traditional semantic segmentation methods. However, Fully Convolutional Network cannot fully utilize boundary information laid in low level features. Thus, Ronneberger \etal~\cite{unet} propose UNet for segmentation which can concatenate feature maps of different scales based on skip connections between encoder and decoder. To further enhance segmentation performance for objects of varying sizes, the Pooling Module~\cite{ppm} and the Atrous Spatial Pyramid Pooling~\cite{deeplab} are introduced, which can extract multi-scale information from a feature map. Later, Attention-based models~\cite{danet,huang2019ccnet} are designed, which can incorporate global semantic information into a feature map. The above-mentioned methods enhance the performance of RGB-based deep learning semantic segmentation from different aspects. 

\subsection{RGB-T Semantic Segmentation}

Similarly, deep learning methods are widely adopted for RGB-T semantic segmentation. However, different from RGB-based Semantic Segmentation, how to fuse deep complementary features from RGB and thermal images effectively is a key problem for RGB-T semantic segmentation. 

Early methods employ addition~\cite{RTFNet,fuseseg} or concatenation~\cite{MFNet,pstnet} operations to directly fuse deep features of RGB and thermal images generated by feature extraction networks. To obtain better segmentation performance, these methods focus on different level features generation for fusion in network architecture. For instance, MFNet~\cite{MFNet} generates different level features using two separate symmetric encoders for RGB and thermal images respectively and fuse them in a shared decoder. RTFNet~\cite{RTFNet} adopts heavyweight backbone ResNet-152s~\cite{resnet} to extract modality features, and fuses different level thermal features with RGB features in encoder stage. FuseSeg~\cite{fuseseg} employs DenseNet-161s~\cite{densenet} as backbone, and fuses the modality features in both encoder and decoder stages. PSTNet~\cite{pstnet} introduces a dual-stream CNN to first extract the RGB semantic information, and then concatenate it to the thermal image to predict the final dense map. The above fusion strategies have been roughly divided into three categories~\cite{li2022rgb}, including decoder fusion~\cite{MFNet}, encoder fusion~\cite{RTFNet,guo2021robust,FEANet,fuseseg,MFFENet} and feature fusion \cite{zhou2021gmnet,EGFNet,liu2022gcnet}. However, recent studies point out that early fusion strategies based on addition or concatenation can easily introduce modality noise while incorporating complementary information~\cite{chen2020bi}. What's more, these methods are facing bottlenecks and cannot achieve significantly-higher segmentation performance.

\begin{figure*}[!ht]
	\centering
	\includegraphics[width=175mm]{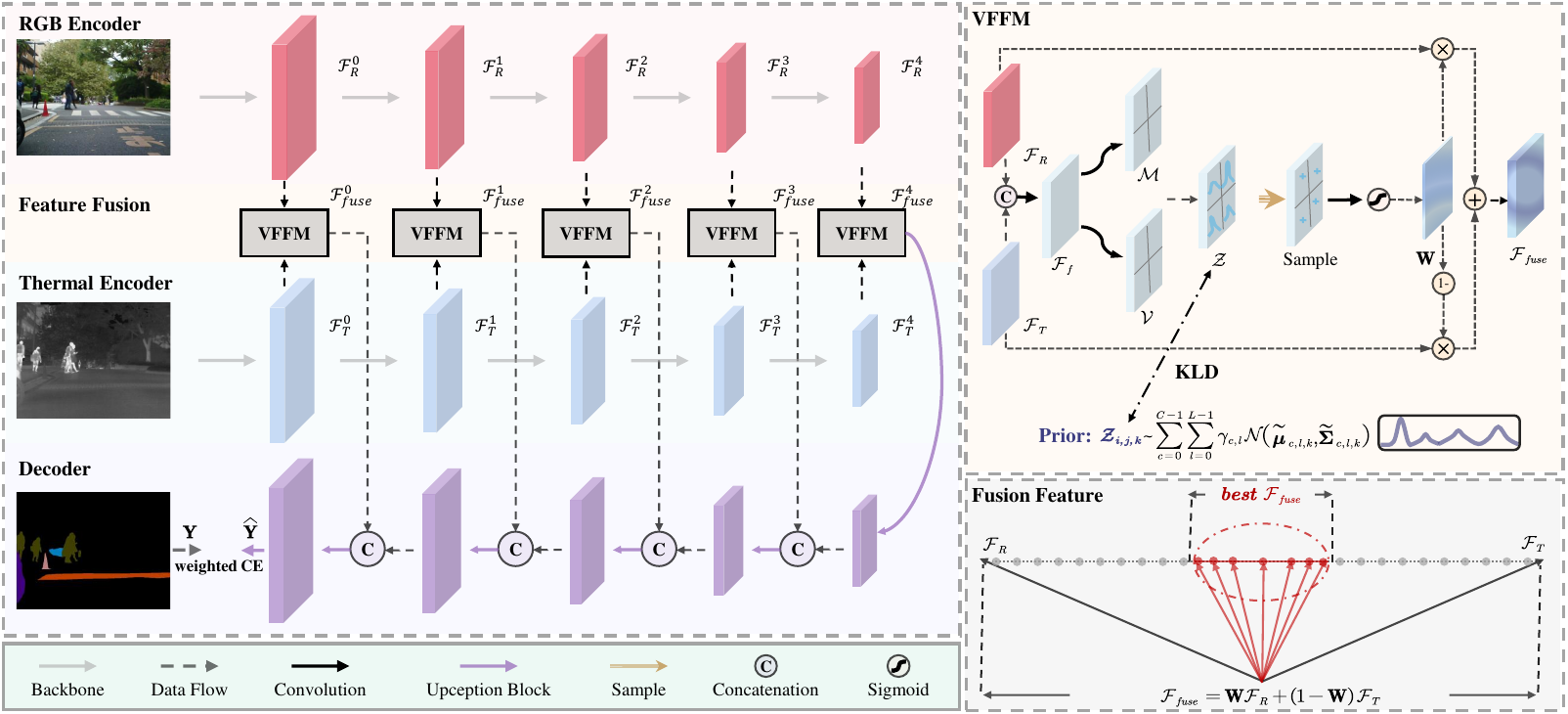}
	\caption{Overall architecture of the proposed VPFNet.}
	\label{fig:architecture}
\end{figure*}

To overcome the bottlenecks caused by fusion strategies based on addition or concatenation, two kinds of solutions are proposed, including designing specific attention-mechanisms-based feature fusion modules~\cite{FEANet,doodlenet,ABMDRNet}, and introducing prior knowledge to supervise fusion models~\cite{zhou2021gmnet,EGFNet,MFFENet}. For the first kind of solutions, FEANet~\cite{FEANet} employs FEAM, which sequentially uses channel and spatial attentions to extract complementary information and suppress modality noise from each single modality before fusing the features.
ABMDRNet~\cite{ABMDRNet} introduces a MSC module and a MCC module to integrate cross-modal multiscale contextual information and their long-range dependencies along the spatial and channel dimensions respectively. DooDleNet~\cite{doodlenet} adopts a spatial attention module based on two rough confidence maps generated by RGB and thermal images respectively to deal with feature disagreement. 
For the second kind of solutions, GMNet~\cite{zhou2021gmnet} introduces boundary label, binary label, and class label to regularize different level fusion features of single modality. Similarly, MFFENet~\cite{MFFENet} introduces the same labels information to regularize fusion features. EGFNet~\cite{EGFNet} employs edge information to guide fusion features generation. The above fusion strategies achieve better segmentation performance compared with early fusion strategies. However, they try to find an optimal fusion feature for the best segmentation, leading to sensitivity to modality noise, class-imbalance, and modality bias. 


\section{Proposed Method}
\label{sec:Method}
In this section, we first provide the overall architecture of VPFNet in Sec.\ref{sec:Method:Architecture}, then, introduce the core module of VPFNet, i.e., the \textit{Variational Feature Fusion Module} (VFFM) in Sec.\ref{Module}. Finally, we provide the loss function in Sec.\ref{Loss}. 


\subsection{Overall Architecture} \label{sec:Method:Architecture}
The overall architecture of the proposed VPFNet for RGB-T semantic segmentation is shown in in Fig.\ref{fig:architecture}, which consists of three parts, including the encoder part, the feature fusion part, and the decoder part. 
For the encoder part, we adopt two separate symmetric ResNet-50s\cite{resnet} as backbones to extract the multi-level discriminative features from RGB and thermal images respectively. The pooling and the fully connected layer of ResNet-50s are removed to preserve spatial resolution. Then, we denote the extracted multi-level features of RGB and thermal images as $ \left\{ \mathcal{F}_{R}^{i}\mid i=0,1,2,3,4 \right\} $ and $ \left\{ \mathcal{F}_{T}^{i}\mid i=0,1,2,3,4 \right\} $ respectively. 
For the feature fusion part, in order to improve representation and generalization abilities, we propose a novel fusion module, VFFM, which regards fusion features as random variables. Using multiple VFFMs, we fuse multi-level features to obtain random samples of multi-level fusion features $ \left\{ \mathcal{F}_{fuse}^{i}\mid i=0,1,2,3,4 \right\} $. More details about VFFM will be introduced in Sec.\ref{Module}.
For the decoder part, we combine multi-level fusion features generated by the feature fusion part to obtain the segmentation result using skip connections and Upception blocks proposed by \cite{RTFNet}. Owing to high computational overhead, we remove two sequential $1\times1$ convolutions from the Upception blocks. As fusion features are random variables, we can obtain $N_s$ segmentation results under $N_s$ samples of fusion features, and obtain the final segmentation result $\mathbf{\overline{Y}}$ by averaging all confidence maps of $N_s$ segmentation results to reduce sensitivity to modality noise.


\subsection{Variational Feature Fusion Module} \label{Module}


To improve generalization and representation abilities, we assume that the optimal features for semantic segmentation are a series of fusion features within a certain range. This assumption is reasonable: First, it can alleviate the sensitivity of fusion features to class-imbalance, modality bias, and modality noise, since the decoder part is required to turn different samples of fusion features to the best segmentation under this assumption. Second, the fusion module based on this assumption can support different biased fusion features, since different RGB-biased and thermal-biased fusion features can be regarded as samples drawn from the fusion feature distribution as shown in Fig.\ref{fig:sample}.

Based on the above assumption, we regard fusion features as random variables. Then, we propose a novel feature fusion module, the \textit{Variational Feature Fusion Module} (VFFM), to fuse different modality features and generate random samples of a fusion feature, as shown in Fig.\ref{fig:architecture}. 
In VFFM, the fusion feature is represented as:
\begin{equation}\label{eq:fusionmodel}
	\left[\mathcal{F}_{fuse}\right]_{i,j,k} = \mathbf{W}_{i,j}\left[\mathcal{F}_R\right]_{i,j,k}+\left( 1-\mathbf{W}_{i,j} \right) \left[\mathcal{F}_T\right]_{i,j,k} 
\end{equation}
where $\mathbf{W}$ is a spatial attention map and $\mathbf{W}_{i,j}$ denotes the attention value at the spatial coordinate $(i,j)$ satisfying $0\leqslant \mathbf{W}_{i,j} \leqslant 1$. When $\mathbf{W}_{i,j}=1$, $\mathcal{F}_{fuse}$ denotes RGB feature; when $\mathbf{W}_{i,j}=0$, $\mathcal{F}_{fuse}$ denotes thermal feature, as shown at bottom right corner of Fig.\ref{fig:architecture}. We also name $\mathbf{W}_{i,j}$ as a \textit{fusion factor}, since it can affect the bias of feature fusion. To obtain different fusion results of $\mathcal{F}_{fuse}$, we regard $\mathbf{W}_{i,j}$ as a random variable. Then, the key problem of VFFM is how to design a neural network to obtain the random samples of $\mathbf{W}_{i,j}$. 

Inspired by VAE~\cite{VAE} and DKPNet~\cite{chen2021variational}, we introduce a latent random 3D-tensor variable $\mathcal{Z}$ to generate the samples of $\mathbf{W}$ instead of directly modeling the distribution function of $\mathbf{W}$. Elements of $\mathcal{Z}$ are assumed to be independently drawn from isotropic Gaussian distributions, Then, we design a network to calculate the means and variances of Gaussian distributions of $\mathcal{Z}$, and generate $\mathbf{W}$ based on random samples of $\mathcal{Z}$. The network design is shown at the top right corner of Fig.\ref{fig:architecture}. 

In VFFM, we first calculate the complementary intermediate fusion feature $\mathcal{F}_f$ as the following:
\begin{equation}\label{eq:Ff}
	\mathcal{F}_f=\mathrm{\sigma}_L\left( \mathrm{BN}\left( \mathrm{Conv}_s\left( \mathrm{Cat}\left( \mathcal{F}_R,\mathcal{F}_T \right);R/r \right) \right) \right) 
\end{equation}
where $\mathrm{Cat}$ denotes concatenation, $R$ denotes the number of channels of $\mathrm{Cat}\left( \mathcal{F}_R,\mathcal{F}_T \right)$, $r$ denotes the squeeze ratio, $\mathrm{Conv}_s(*;R/r)$ denotes convolution with kernel size $s$ and $R/r$ output channels, $\mathrm{BN}$ denotes Batchnorm, and $\mathrm{\sigma}_L$ denotes LeakyReLU activate function with rate $0.2$ , which can enhance the feature representation ability \cite{MFNet}. Then, we use $\mathcal{F}_f$ to generate the mean $\mathcal{M}_{i,j,k}$ and variance $\mathcal{V}_{i,j,k}$ of isotropic Gaussian distribution of $\mathcal{Z}_{i,j,k}$, i.e.,
\begin{subequations}\label{eq:Zdis}
	\begin{gather} 
		\mathcal{Z}_{i,j,k}\sim \mathcal{N}(\mathcal{M}_{i,j,k},\mathcal{V}_{i,j,k}) \\
		\mathcal{M}=  \mathrm{Conv}_1\left( \mathcal{F}_f; d \right),~\log \mathcal{V} =\mathrm{Conv}_1\left( \mathcal{F}_f; d \right) 
	\end{gather}
\end{subequations}
where $\mathrm{Conv}_1(*;d)$ denotes a $1\times 1$ convolutional block with $d$ output channels. Finally, $\mathbf{W}$ can be obtained by:
\begin{equation}\label{eq:Wgen}
	\mathbf{W}=\mathrm{\sigma_S}\left( \mathrm{Conv}_1\left(  \mathcal{Z}; 1 \right) \right)
\end{equation}
where $\text{Conv}_1(*;1)$ denotes $ 1\times1 $ convolution with only one output channel, elements of $\mathcal{Z}$ are drawn from the distribution of Eqn.(\ref{eq:Zdis}). $\mathrm{\sigma_S}$ denotes Sigmoid activation function to normalize elements of $\mathbf{W}$ between 0 and 1. Based on the above network and Eqn.(\ref{eq:fusionmodel}), we can obtain the fusion feature sample of $\mathcal{F}_{fuse}$.




\subsection{Loss Function} \label{Loss}

Theoretically, we can directly employ the cross-entropy loss as the RGB-T semantic segmentation loss. Using this loss function, the fusion network will be trained to obtain the best segmentation under different fusion feature samples of $\mathcal{F}_{fuse}$ obtained from a pair of RGB and thermal images, thus, it is robust to modality noise. However, the cross-entropy loss does not take data class-imbalance~\cite{imbalancereview} into consideration. Moreover, recent studies point out that uniform objective designed for all modalities could lead to under-optimized uni-modal representations\cite{peng2022balanced}, i.e., modality bias.

To avoid class-imbalance and modality bias, we employ the weighted cross-entropy loss and introduce prior information (\eg, illumination and category) to control the fusion factor distribution $p\left( \mathbf{W} \right)$. Specifically, inspired by DKPNet~\cite{chen2021variational}, we maximize the log-likelihood of the conditional probability $p\left( \mathbf{W}\mid \mathcal{F}_R,\mathcal{F}_T,\mathbf{Y}, l \right)$ as follows:
\begin{equation}\label{eq:ELBO}
	\begin{aligned} 
		&\log \left( p\left( \mathbf{W}\mid \mathcal{F}_R,\mathcal{F}_T,\mathbf{Y},l \right) \right)\\ 
		&= \log \int{q\left( \mathcal{Z}\mid \mathcal{F}_R,\mathcal{F}_T,\mathbf{Y},l \right) \frac{p\left( \mathbf{W},\mathcal{Z}\mid \mathcal{F}_R,\mathcal{F}_T,\mathbf{Y},l \right)}{q\left( \mathcal{Z}\mid \mathcal{F}_R,\mathcal{F}_T,\mathbf{Y},l \right)}dz} 
		\\
		&\geqslant \mathbb{E} _{q\left( \mathcal{Z}\mid \mathcal{F}_R,\mathcal{F}_T,\mathbf{Y},l \right)}\left[ \log \left( p\left( \mathbf{W}\mid \mathcal{F}_R,\mathcal{F}_T,\mathbf{Y},\mathcal{Z},l \right) \right) \right] \\
		&\qquad -KL\left[ q\left( \mathcal{Z}\mid \mathcal{F}_R,\mathcal{F}_T,\mathbf{Y},l \right) \parallel p\left( \mathcal{Z}\mid \mathcal{F}_R,\mathcal{F}_T,\mathbf{Y},l \right) \right]
	\end{aligned}
\end{equation}
where $l$ denotes illumination scenario prior (\eg, day and night) of the input images, $\mathbf{Y}$ denotes segmentation containing category prior of every pixel, $q\left( \mathcal{Z}\mid \mathcal{F}_R,\mathcal{F}_T,\mathbf{Y},l \right)$ denotes the variational posterior distribution of Eqn.(\ref{eq:Zdis}), $p\left( \mathcal{Z}\mid \mathcal{F}_R,\mathcal{F}_T,\mathbf{Y},l \right)$ denotes the prior distribution of $\mathbf{Z}$. 

According to \cite{VAE}, we can maximize the evidence lower bound (ELBO) as a proxy for maximizing the log-likelihood. In Eqn(\ref{eq:ELBO}), ELBO includes two terms.
The first term $\mathbb{E} _{q}\left[ \log \left( p\left( \mathbf{W}\mid \mathcal{F}_R,\mathcal{F}_T,\mathbf{Y},\mathcal{Z},l \right) \right) \right]$ aims to generate the best fusion factor matrix $\mathbf{W}$ from latent variables $\mathcal{Z}$ to improve segmentation performance, i.e., it corresponds to segmentation estimation loss. Considering data class-imbalance~\cite{imbalancereview}, we describe this term as the weighted cross-entropy loss between the ground-truth segmentation $\mathbf{Y}$ and the estimated segmentation $\mathbf{\hat{Y}(\mathbf{W})}$ based on the fusion factor matrix $\mathbf{W}$. 
The second term is the KL distance between the posterior distribution $q\left( \mathcal{Z}\mid \mathcal{F}_R,\mathcal{F}_T,\mathbf{Y},l \right)$ and the prior distribution $p\left( \mathcal{Z}\mid \mathcal{F}_R,\mathcal{F}_T,\mathbf{Y},l \right)$. This term can indirectly control the posterior distribution of fusion factor distribution $p\left( \mathbf{W} \right)$ to alleviate modality bias~\cite{chen2021variational} based on illumination and category prior information, since $\mathbf{W}$ is obtained from $\mathcal{Z}$ based on Eqn.(\ref{eq:Wgen}). 

To design the prior distribution $p\left( \mathcal{Z}\mid \mathcal{F}_R,\mathcal{F}_T,\mathbf{Y},l \right)$, we assume that each element of $\mathcal{Z}$ is independently drawn from Gaussian mixture distribution with $L \times C$ Gaussian components, and 
\begin{subequations}\label{eq:prior}
	\begin{align} 
		&p\left( \mathcal{Z}_{i,j,k}\mid \mathcal{F}_R,\mathcal{F}_T,\mathbf{Y}_{i,j}=c,l \right) \notag\\
		&~~\approx p\left( \mathcal{Z}_{i,j,k}\mid \mathbf{Y}_{i,j}=c,l \right)=\mathcal{N} \left( \widetilde{\bm{\mu}}_{c,l,k}, \widetilde{\bm{\Sigma}}_{c,l,k} \right)\\
		&p(\mathbf{Y}_{i,j}=c,l) =\gamma _{c,l}\geq 0,~\sum_{c=0}^{C-1}{\sum_{l=0}^{L-1}{\gamma _{c,l}}}=1\\
		&p\left( \mathcal{Z}_{i,j,k} \right)\approx \sum_{c=0}^{C-1}{\sum_{l=0}^{L-1}{\gamma _{c,l}\mathcal{N} \left( \widetilde{\bm{\mu}}_{c,l,k}, \widetilde{\bm{\Sigma}} _{c,l,k} \right)}},
	\end{align}
\end{subequations}
where $C$ denotes the number of categories, $L$ denotes the number of illumination conditions, $\widetilde{\bm{\mu}}_{c,l,k}$ and $\widetilde{\bm{\Sigma}}_{c,l,k}$ represent the mean and variance with different category and illumination condition respectively. In this paper, we only consider two illumination conditions, i.e., day and night, thus, $L=2$. For convenience, we set $\gamma _{c,l}=\frac{1}{LC}$. As for $\widetilde{\bm{\mu}}_{c,l,k}$ and $\widetilde{\bm{\Sigma}}_{c,l,k}$, these parameters are set to be learnable during the training process, thus, prior of $\mathcal{Z}$ can record the global statistical information of the training dataset. 

Then, the KL distance in Eqn.(\ref{eq:ELBO}) can be formulated as:
\begin{equation}
	\label{eq:KLdis}
	\begin{split} 
		&KL\left[ q\left( \mathcal{Z}\mid \mathcal{F}_R,\mathcal{F}_T,\mathbf{Y},l \right) \parallel p\left( \mathcal{Z}\mid \mathcal{F}_R,\mathcal{F}_T,\mathbf{Y},l \right) \right] \\
		&=\frac{1}{D}\sum_{i,j,k}\frac{1}{2}\left[ \log \frac{ \widetilde{\bm{\Sigma}}_{c,l,k} }{  \mathcal{V}_{i,j,k} }+ tr\left( \widetilde{\bm{\Sigma}}_{c,l,k}^{-1}\mathcal{V}_{i,j,k} \right) -d\right.  \\
		&\left. \qquad+\left( \widetilde{\bm{\mu}}_{c,l,k}-\mathcal{M}_{i,j,k} \right) \widetilde{\bm{\Sigma}}_{c,l,k}^{-1}\left( \widetilde{\bm{\mu}}_{c,l,k}-\mathcal{M}_{i,j,k} \right) ^T \right] 
	\end{split}
\end{equation}
where $d$ denotes the channel dimension of $\mathcal{Z}$, $D$ denotes the number of elements in $\mathcal{Z}$, $l$ denotes the illumination condition of each pair of RGB and thermal images, $c$ denotes the pixel category label which is related to the segmentation mask of $\mathcal{Z}$ at the spatial coordinate $(i,j)$. Since the spatial sizes of $\mathcal{Z}$ and $\mathbf{Y}$ are different, we resize the segmentation mask $\mathbf{Y}$ using the nearest neighbor method to obtain the segmentation mask of $\mathcal{Z}$.


Based on the above modeling, the loss function of the proposed VPFNet can be formulated as:
\begin{equation}\label{eq:loss}
	\begin{aligned} 
		L_{total}=\frac{1}{N_b}\sum_{b=1}^{N_b}{ \left(\mathrm{WeightCE}\left( \mathbf{\hat{Y}}_b,\mathbf{Y}_b \right) + \beta \overline{KL}_b\right) }
	\end{aligned}
\end{equation}
where $\mathrm{WeightCE}(*,*)$ denotes the weighted cross-entropy loss between ground-truth label $\mathbf{Y}$ and segmentation prediction $\mathbf{\hat{Y}}$ following ~\cite{paszke2016enet}, $\overline{KL}$ denotes the average KL distance of Eqn.(\ref{eq:KLdis}) for all VFFMs used to generate multi-level features, $N_b$ denotes the number of RGB-thermal image pairs in the training mini-batch. $\beta$ is a hyper-parameter to avoid posterior collapse following \cite{betaVAE}. 


\section{Experiments}

\begin{table}[!tp]
	\setlength{\abovecaptionskip}{0pt}
	\caption{The results of VPFNet with different values of $\beta$.}
	\label{tbl:ablation-beta}
	\centering
	\setlength{\tabcolsep}{1.32em}
	\renewcommand\arraystretch{0.6}
	\begin{tabular}{c|ccccc}
		\toprule
		$\beta$ & 0     & 0.3 & 0.5   & 0.7   & 1 \\ \midrule
		mIoU & 56.78 & 57.09                      & \textbf{57.61} & 56.71 & 56.35                    \\ \bottomrule
	\end{tabular}
\end{table}

\begin{table}[!tp]
	\setlength{\abovecaptionskip}{0pt}
	\newcommand{\tabincell}[2]{\begin{tabular}{@{}#1@{}}#2\end{tabular}}
	\caption{The results of probabilistic and non-probabilistic fusion models}
	\label{tbl:ablation-prob}
	\centering
	\setlength{\tabcolsep}{0.44em}
	\renewcommand\arraystretch{0.8}
	\begin{tabular}{c|c|c|c}
		\toprule
		Methods            & \tabincell{c}{non-prob. fusion\\without attention}    & \tabincell{c}{non-prob. fusion\\with attention}       & \tabincell{c}{probabilistic fusion \\(VPFNet without KL)}     \\ \midrule
		mAcc               & 69.96                                & 68.70                                & \textbf{71.08}  \\ \midrule
		mIoU               & 55.99                                & 56.35                                & \textbf{56.78}   \\\bottomrule
	\end{tabular}
\end{table}


\begin{table}[!tp]
	\setlength{\abovecaptionskip}{0pt}
	\caption{{The results of VPFNet with different loss functions}}
	\label{tbl:ablation-weight}
	\setlength{\tabcolsep}{0.69em}
	\centering
	\renewcommand\arraystretch{0.8}
	\begin{tabular}{c|cc|cc|cc}
		\toprule
		\multirow{2}{*}{Loss Func.}  & \multicolumn{2}{c|}{Car stop} & \multicolumn{2}{c|}{Guardrail} & \multirow{2}{*}{mAcc} & \multirow{2}{*}{mIoU} \\ \cmidrule(lr){2-5}
		& Acc          & IoU         & Acc           & IoU          &                       &                       \\ \midrule
		CE                          &  32.41         & 27.98        & 20.87          & 3.91          & 65.61                 & 55.56                 \\
		CE+KL                       &  32.87         & 28.48        & 29.35          & 4.77          & 67.96                 & 56.50                 \\
		WeightCE                    &  34.82         & 28.66        & 52.39          & 7.48          & \textbf{71.08}                 & 56.78                 \\
		WeightCE + KL               &  38.33         & 31.67        & 46.32          & 9.40          & 70.91                 & \textbf{57.61}                 \\ \bottomrule
	\end{tabular}
\end{table}

\begin{table}[!tp]
	\setlength{\abovecaptionskip}{0pt}
	\newcommand{\tabincell}[2]{\begin{tabular}{@{}#1@{}}#2\end{tabular}}
	\caption{The results of VPFNet with different KL prior regularization}
	\label{tbl:ablation-prior}
	\centering
	\setlength{\tabcolsep}{1.28em}
	\renewcommand\arraystretch{0.8}
	\begin{tabular}{c|p{0.9cm}<{\centering}p{0.9cm}<{\centering}|cc}
		\toprule
		Loss Func.                           & Illumination                     & Category                         & mAcc  & mIoU  \\ \midrule
		without KL & -                                & -                                & \textbf{71.08} & 56.78 \\ \midrule
		\multirow{3}{*}{\tabincell{c}{KL with \\different prior\\information}}                            & { \checkmark} &                                  & 69.81 & 56.61 \\
		&                                  & { \checkmark} & 69.51 & 56.88 \\
		& { \checkmark} & { \checkmark} & 70.91 & \textbf{57.61} \\ \bottomrule
	\end{tabular}
\end{table}


\begin{table}[!tp]
	\setlength{\abovecaptionskip}{0pt}
	\newcommand{\tabincell}[2]{\begin{tabular}{@{}#1@{}}#2\end{tabular}}
	\caption{The results of VPFNet under different values of $N_s$}
	\label{tbl:ablation-ns}
	\centering
	\setlength{\tabcolsep}{1.31em}
	\renewcommand\arraystretch{0.6}
	\begin{threeparttable}
		\begin{tabular}{c|c|c|c|c|c}
			\toprule
			$N_s$& 1*     & 5           & 10          & 20          & 50          \\ \midrule
			mAcc & 70.91 & 69.75 & 69.72 & 69.77 & 69.82 \\\midrule
			mIoU & 57.61 & 57.62 & 57.65 & 57.71 & 57.72 \\ \bottomrule
		\end{tabular}
		\begin{tablenotes}
			\scriptsize
			\item *$N_s=1$ means directly adopting the mean of latent variable $\mathcal{Z}$ to generate fusion feature.
		\end{tablenotes}
	\end{threeparttable}
\end{table}

\subsection{Experiment Setup}

\subsubsection{Datasets}
We conduct experiments on MFNet~\cite{MFNet} dataset and PST900~\cite{pstnet} dataset, which are the only two public datasets for RGB-T semantic segmentation. For MFNet dataset, it contains 1569 pairs of annotated RGB and thermal images with 8 object categories in addition to the background category, including 820 pairs of images captured at daytime and 749 pairs of images captured at nighttime. Both RGB and thermal images have the same spatial resolution of 480$\times$640. For fairness, we directly adopt the dataset splitting scheme following ~\cite{MFNet}. For PST900 dataset, it consists of 894 pairs aligned RGB and thermal images captured at nighttime with 4 object categories in addition to the background category. Following \cite{EGFNet}, we resize the resolution of image to 640 $\times$ 1280. Then, we adopt the same training, testing and verification setting as in~\cite{pstnet}.  

\begin{figure}[!tp]
	\centering
	\includegraphics[width=80mm]{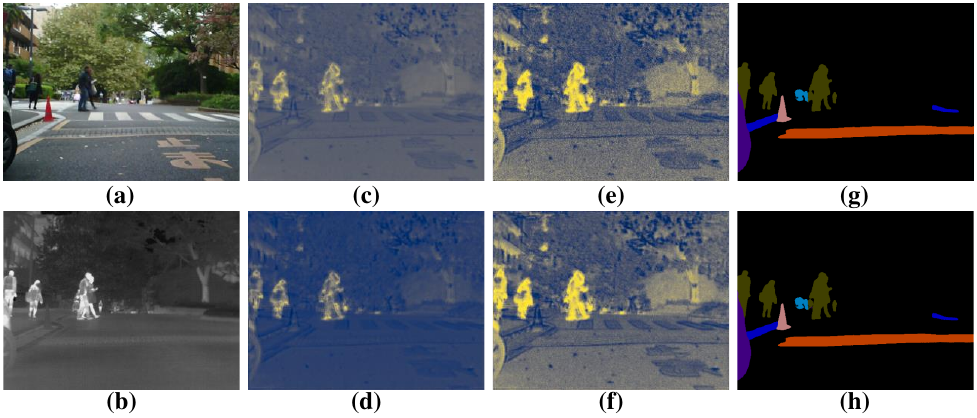}
	\caption{Visualization example for the probabilistic fusion model. (a)-(h) RGB image, thermal image, the mean of latent variable $\mathcal{Z}$, the variance of latent variable $\mathcal{Z}$, a random sample of fusion feature, the average of five fusion feature samples, the segmentation result under a fusion feature sample, the average segmentation result under five fusion feature samples.}
	\label{fig:probabilistic}
\end{figure}


\begin{table*}[!tp]
	\footnotesize
	\setlength{\abovecaptionskip}{0pt}
	\caption{Results on the MFNet dataset.  '-' denotes that the corresponding results are missed in~\cite{pstnet}. For RGB-based methods, 3c means the input images are RGB images and 4c means the input images are 4 channels RGB-T images. For RGB-D methods, we replace the depth images with thermal images.}
	\label{tbl:MFNet-test}   
	\resizebox{\textwidth}{!}{
		\begin{tabular}{ccccccccccccccccccccc}
			\toprule
			\multirow{2}{*}{Methods} & \multirow{2}{*}{Type} & \multirow{2}{*}{Year} & \multicolumn{2}{c}{Car}         & \multicolumn{2}{c}{Person}      & \multicolumn{2}{c}{Bike}        & \multicolumn{2}{c}{Curve}       & \multicolumn{2}{c}{Car Stop}    & \multicolumn{2}{c}{Guardrail}   & \multicolumn{2}{c}{Color Cone}  & \multicolumn{2}{c}{Bump}        & \multirow{2}{*}{mAcc} & \multirow{2}{*}{mIoU} \\ \cmidrule(lr){4-19}
			&                       &                       & Acc            & IoU            & Acc            & IoU            & Acc            & IoU            & Acc            & IoU            & Acc            & IoU            & Acc            & IoU            & Acc            & IoU            & Acc            & IoU            &                       &                       \\ \midrule
			CCNet-3c                 & RGB                   & ICCV 2019             & 87.92          & 80.28          & 56.56          & 45.82          & 74.33          & 58.13          & 34.30          & 26.78          & 32.96          & 25.19          & 26.64          & 4.49           & 41.92          & 36.08          & 46.19          & 36.22          & 55.52                 & 45.56                 \\
			CCNet-4c                 & RGB                   & ICCV 2019             & 88.07          & 81.97          & 61.09          & 51.88          & 74.35          & 59.53          & 46.83          & 36.27          & 28.06          & 22.71          & 2.84           & 0.55           & 43.37          & 37.07          & 51.14          & 44.19          & 54.98                 & 47.95                 \\
			DANet-3c                 & RGB                   & CVPR 2019             & 88.83          & 80.07          & 54.63          & 44.31          & 72.04          & 57.38          & 27.66          & 22.30          & 33.86          & 25.89          & 22.70          & 4.21           & 37.19          & 31.73          & 46.42          & 35.28          & 53.57                 & 44.24                 \\
			DANet-4c                 & RGB                   & CVPR 2019             & 85.73          & 80.48          & 58.69          & 49.63          & 70.22          & 57.14          & 45.11          & 35.38          & 27.42          & 22.57          & 29.80          & 5.93           & 36.08          & 32.34          & 46.74          & 38.01          & 55.43                 & 46.53                 \\ \midrule
			FuseNet                  & RGBD                  & ACCV 2016             & 81.00          & 75.60          & 75.20          & 66.30          & 64.50          & 51.90          & 51.00          & 37.80          & 17.40          & 15.00          & 0.00           & 0.00           & 31.10          & 21.40          & 51.90          & 45.00          & 52.40                 & 45.60                 \\
			D-CNN                    & RGBD                  & ECCV 2018             & 85.20          & 77.00          & 61.70          & 53.40          & 76.00          & 56.50          & 40.20          & 30.90          & 41.30          & 29.30          & 22.80          & 8.50           & 32.90          & 30.10          & 36.50          & 32.30          & 55.10                 & 46.10                 \\
			SAGate                   & RGBD                  & ECCV2020              & 86.00          & 73.80          & 80.80          & 59.20          & 69.40          & 51.30          & 56.70          & 38.40          & 24.70          & 19.30          & 0.00           & 0.00           & 56.90          & 24.50          & 52.10          & 48.80          & 58.30                 & 45.80                 \\ \midrule
			MFNet                    & RGBT                  & IROS 2017             & 77.20          & 65.90          & 67.00          & 58.90          & 53.90          & 42.90          & 36.20          & 29.90          & 12.50          & 9.90           & 0.10           & 0.00           & 30.30          & 25.20          & 30.00          & 27.70          & 45.10                 & 39.70                 \\
			RTFNet-50                & RGBT                  & RA-L 2019             & 91.30          & 86.30          & 78.20          & 67.80          & 71.50          & 58.20          & 69.80          & 43.70          & 32.10          & 24.30          & 13.40          & 3.60           & 40.40          & 26.00          & 73.50          & \textbf{57.20}          & 62.20                 & 51.70                 \\
			RTFNet-152               & RGBT                  & RA-L 2019             & 93.00          & 87.40          & 79.30          & 70.30          & 76.80          & 62.70          & 60.70          & 45.30          & 38.50          & 29.80          & 0.00           & 0.00           & 45.50          & 29.10          & 74.70          & 55.70          & 63.10                 & 53.20                 \\
			PSTNet                   & RGBT                  & ICRA 2020             & -              & 76.80          & -              & 52.60          & -              & 55.30          & -              & 29.60          & -              & 25.10          & -              & \textbf{15.10} & -              & 39.40          & -              & 45.00          & -                     & 48.40                 \\
			FuseSeg-161              & RGBT                  & T-ASE 2021            & 93.10          & \textbf{87.90} & 81.40          & 71.70          & 78.50          & \textbf{64.60} & 68.40          & 44.80          & 29.10          & 22.70          & 63.70          & 6.40           & 55.80          & 46.90          & 66.40          & 47.90          & 70.60                 & 54.50                 \\
			FEANet                   & RGBT                  & IROS 2021             & 93.30          & 87.80          & 82.70          & 71.10          & 76.70          & 61.10          & 65.50          & 46.50          & 26.60          & 22.10          & \textbf{70.80} & 6.60           & \textbf{66.60} & \textbf{55.30} & \textbf{77.30} & 48.90          & 73.20                 & 55.30                 \\
			ABMDRNet                 & RGBT                  & CVPR 2021             & 94.30          & 84.80          & 90.00          & 69.60          & 75.70          & 60.30          & 64.00          & 45.10          & 44.10          & 33.10          & 31.00          & 5.10           & 61.70          & 47.40          & 66.20          & 50.00          & 69.50                 & 54.80                 \\
			EGFNet                   & RGBT                  & AAAI 2022             & \textbf{95.80} & 87.60          & 89.00          & 69.80          & \textbf{80.60} & 58.80          & 71.50          & 42.80          & \textbf{48.70} & 33.80          & 33.60          & 7.00           & 65.30          & 48.30          & 71.10          & 47.10          & 72.70                 & 54.80                 \\
			MDRNet+                  & RGBT                  & TNNLS 2023            & 95.20          & 87.10          & \textbf{92.50} & 69.80          & 76.20          & 60.90          & \textbf{72.00} & 47.80          & 42.30          & \textbf{34.20} & 66.80          & 8.20           & 64.80          & 50.20          & 63.50          & 55.00          & \textbf{74.70}        & 56.80                 \\ \midrule
			Ours                     & RGBT                  & -                     & 93.43          & 87.10          & 87.70          & \textbf{73.97} & 73.82          & 62.57          & 68.84          & \textbf{48.15} & 38.33          & 31.67          & 46.32          & 9.40           & 60.06          & 52.25          & 70.69          & 55.31          & 70.91                 & \textbf{57.61}        \\ \bottomrule
	\end{tabular}}
\end{table*}

\begin{figure*}[!tp]
	\centering
	\setlength{\abovecaptionskip}{5pt}
	\includegraphics[width=170mm]{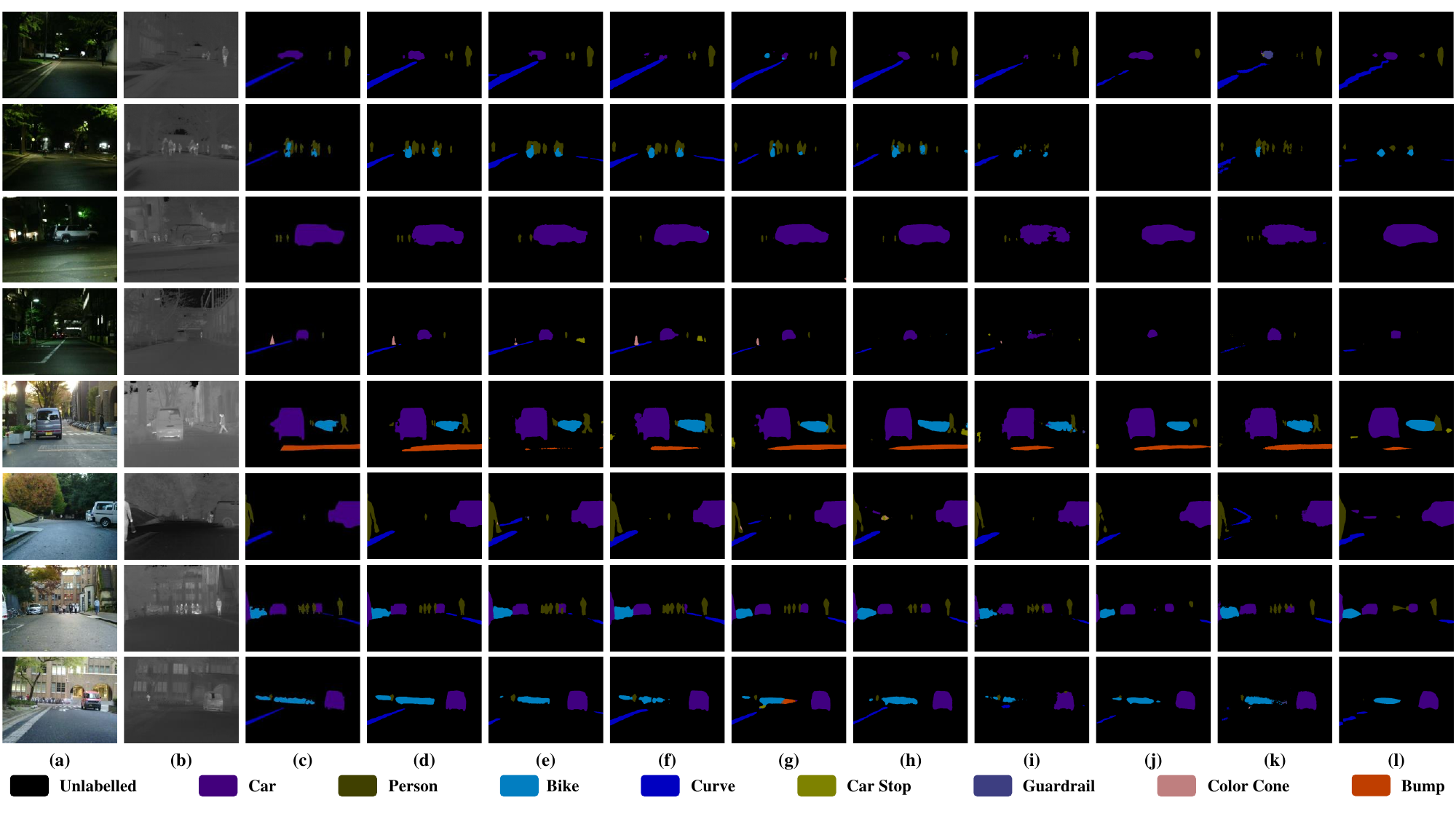}
	\caption{Visual comparisons of different methods. (a) RGB images; (b) Thermal images; (c) Ground Truth; (d) Ours; (e) MDRNet;
		(f) EGFNet; (g) FEANet; (h) RTFNet; (i) MFNet; (j) SA-Gate; (k) FuseNet; (l) CCNet.}
	\label{fig:results}
\end{figure*}

\subsubsection{Evaluation Metrics}
Mean accuracy (mAcc) and mean intersection over union (mIoU) are adopted to evaluate the performance of different semantic segmentation models following previous related works~\cite{MFNet,RTFNet,EGFNet,ABMDRNet}.

\subsubsection{Implementation Details}
The proposed VPFNet is implemented by PyTorch 1.11.0, CUDA 11.3, and Python 3.9.13. For training, the backbone is pretrained on ImageNet~\cite{deng2009imagenet}. Then, following the training strategy in~\cite{EGFNet}, we use random flipping and random cropping for data augmentation, and train the proposed VPFNet to its convergence within 300 epochs using the Ranger optimizer~\cite{wright2021ranger21} with an initial learning rate and weight decay of 5e-5 and 5e-4 respectively. 
The training mini-batch size $N_b$ is set to 3 in Eqn.(\ref{eq:loss}). we run all experiments on a NVIDIA-V100 GPU.

\subsection{Ablation Study}
In this section, we first discuss the selection of hyperparamters in VPFNet. Then, we validate the effectiveness of each modeling component in the proposed VPFNet, including probabilistic fusion, loss function, and KL prior regularization. All experiments in this section are conducted based on MFNet dataset.



\subsubsection{Selection of Hyperparameters}
We first select the value of hyperparamters $s$, $r$, and $d$ in Eqn.(\ref{eq:Ff}) and Eqn.(\ref{eq:Zdis}). We carried out ablation experiments with $s\in\left\{3,5,7\right\}$, $r\in\left\{8,16,32\right\}$ and $d\in\left\{2,4,6,8,10\right\}$, and found VPFNet works best when $s=7$, $r=16$ and $d=8$.

Then, we select the value of weight factor $\beta$ in Eqn.(\ref{eq:loss}). Theoretically, if $\beta$ is too big, the KL prior regularization term in Eqn.(\ref{eq:loss}) will weaken feature representation ability; if $\beta$ is too small, the regularization term will have little influence on segmentation performance. Thus, we carry out ablation experiments with $\beta\in\left\{0, 0.3, 0.5, 0.7, 1\right\}$. The results are shown in Table \ref{tbl:ablation-beta}. From Table \ref{tbl:ablation-beta}, we can find that VPFNet can achieve the best performance when $\beta=0.5$.

Moreover, we select the value of $N_s$ mentioned in Sec.\ref{Architecture}. Theoretically, the segmentation result tends to be stable with the increase of $N_s$. However, $N_s$ is required to be small considering the balance of computational cost and performance. Thus, we carry out ablation experiments with $N_s\in\left\{1, 5, 10, 20, 50\right\}$. The results are shown in Table \ref{tbl:ablation-ns}. From the results, we can find that the mIoU value tends to be stable when $N_s\geq 20$. However, mIoU values at $N_s\geq 20$ are only slightly better than that at $N_s=1$.Thus, for convenience, we approximate the average segmentation result by directly using the mean of $\mathcal{Z}$ to generate the fusion feature sample.

\subsubsection{Effectiveness of Probabilistic Fusion}
To validate the effectiveness of probabilistic fusion, we design probabilistic and non-probabilistic fusion models based on VPFNet for comparison. The comparison results are shown in Table \ref{tbl:ablation-prob}. The non-probabilistic fusion model without attention is designed by replacing VFFM with element-wise addition to fuse modality features in VPFNet; The non-probabilistic fusion model with attention is designed by directly utilizing the mean $\mathcal{M}$ to replace the random sample $\mathcal{Z}$ in Eqn.(\ref{eq:Wgen}) (i.e., deleting the variance $\mathcal{V}$ and the random sampling process); The probabilistic fusion model is VPFNet without KL prior regularization. All these fusion models share the same weighted cross-entropy loss. From Table \ref{tbl:ablation-prob}, we can find that the probabilistic fusion model outperforms two non-probabilistic fusion models under different evaluation metrics, which prove the effectiveness of probabilistic fusion.

\begin{table}[]
	\footnotesize  
	\setlength{\abovecaptionskip}{0pt}
	\caption{Results from daytime and nighttime images.}
	\label{tbl:MFNet-daynight}
	\centering
	\setlength{\tabcolsep}{1.0em}
	\renewcommand\arraystretch{0.9}
	\begin{tabular}{cccccc}
		\toprule
		\multirow{2}{*}{Methods} & \multirow{2}{*}{Type} & \multicolumn{2}{c}{Daytime}     & \multicolumn{2}{c}{Nighttime}   \\ \cmidrule(l){3-6} 
		&                       & mAcc           & mIoU           & mAcc           & mIoU           \\ \midrule
		CCNet-3c                 & RGB                   & 50.26          & 41.70          & 50.30          & 42.72          \\
		CCNet-4c                 & RGB                   & 48.67          & 41.54          & 51.89          & 46.59          \\
		DANet-3c                 & RGB                   & 50.18          & 41.05          & 48.20          & 41.45          \\
		DANet-4c                 & RGB                   & 48.48          & 41.12          & 51.78          & 45.02          \\ \midrule
		FuseNet                  & RGBD                  & 46.71          & 40.13          & 49.11          & 42.02          \\
		D-CNN                    & RGBD                  & 50.60          & 42.40          & 50.70          & 43.20          \\
		SA-Gate                  & RGBD                  & 49.30          & 37.90          & 56.90          & 45.60          \\ \midrule
		MFNet                    & RGBT                  & 42.60          & 36.10          & 41.40          & 36.80          \\
		RTFNet-152               & RGBT                  & 60.00          & 45.80          & 60.70          & 54.80          \\
		FuseSeg-161              & RGBT                  & 62.10          & 47.80          & 67.30          & 54.60          \\
		FEANet                   & RGBT                  & 51.57          & 39.85          & 56.60          & 50.09          \\
		ABMDRNet                 & RGBT                  & 58.40          & 46.70          & 68.30          & 55.50          \\
		EGFNet                   & RGBT                  & \textbf{74.40} & 47.30          & 68.00          & 55.00          \\
		MDRNet+                  & RGBT                  & 59.19          & 47.71          & \textbf{71.95} & 55.60          \\ \midrule
		Ours                     & RGBT                  & 63.29          & \textbf{50.32} & 67.81          & \textbf{56.64} \\ \bottomrule
	\end{tabular}
\end{table}


To further show the advantage of probabilistic fusion, we display a visualization example in Fig.\ref{fig:probabilistic}. From Fig.\ref{fig:probabilistic}, we can find that the fusion feature in Fig.\ref{fig:probabilistic}(e) is noisier than the average feature in Fig.\ref{fig:probabilistic}(f). However, the segmentation result of Fig.\ref{fig:probabilistic}(g) is nearly the same as that of Fig.\ref{fig:probabilistic}(h) despite that it is obtained based on noisy fusion feature, which prove that the segmentation based on probabilistic fusion is robust to noise.

\subsubsection{Effectiveness of Loss Function}
To validate the effectiveness of loss function, we design different loss functions for comparison, as shown in Table \ref{tbl:ablation-weight}. In Table \ref{tbl:ablation-weight}, we additionally provide the results of two categories, Car stop and Guardrail, which have only a few training samples. From the results, we can find that the weight cross entropy (WeightCE) loss can alleviate class-imbalance and enhance the segmentation performance compared with the cross entropy (CE) loss. Besides, the KL prior regularization in Eqn.(\ref{eq:loss}) is necessary, since it can significantly enhance segmentation performance coordinated with the CE or WeightCE loss.

\begin{table}[]
	\footnotesize  
	\setlength{\abovecaptionskip}{0pt}
	\caption{Results when the input images lack a certain modal information. Only RGB and Only Thermal mean the input images are RGB images or thermal images (we set the missing modality to zero when testing).}
	\label{tbl:MFNet-rgninf}
	\centering
	\setlength{\tabcolsep}{0.55em}
	\renewcommand\arraystretch{0.9}
	\begin{tabular}{cccccccc}
		\toprule
		\multirow{2}{*}{Methods} & \multirow{2}{*}{Type} & \multicolumn{2}{c}{Only RGB}    & \multicolumn{2}{c}{Only Thermal} & \multicolumn{2}{c}{Worst Modality} \\ \cmidrule(l){3-8} 
		&                       & mAcc           & mIoU           & mAcc            & mIoU           & mAcc             & mIoU            \\ \midrule
		FuseNet                  & RGBD                  & 11.11          & 10.31          & 41.33           & 36.85          & 11.11            & 10.31           \\
		SAGate                   & RGBD                  & 32.01          & 30.57          & 13.34           & 12.51          & 13.34            & 12.51           \\ \midrule
		MFNet                    & RGBT                  & 26.62          & 24.78          & 19.65           & 16.64          & 19.65            & 16.64           \\
		RTFNet                   & RGBT                  & 44.89          & 37.30          & 26.41           & 24.57          & 26.41            & 24.57           \\
		FEANet                   & RGBT                  & 15.96          & 8.69           & \textbf{58.35}  & \textbf{48.72} & 15.96            & 8.69            \\
		MDRNet                   & RGBT                  & \textbf{57.11} & \textbf{45.89} & 41.98           & 30.19          & 41.98            & 30.19           \\ \midrule
		Ours                     & RGBT                  & 48.14          & 41.08          & 42.20           & 35.80          & \textbf{42.20}            & \textbf{35.80}  \\ \bottomrule
	\end{tabular}
\end{table}


\begin{table*}[]
	\footnotesize  
	\setlength{\abovecaptionskip}{0pt}
	\caption{Results from PST900 dataset~\cite{pstnet}.}
	\label{tbl:PST900}
	\centering
	\setlength{\tabcolsep}{0.92em}
	\renewcommand\arraystretch{0.9}
	\begin{tabular}{cccccccccccccc}
		\toprule
		\multirow{2}{*}{Methods} & \multirow{2}{*}{Type} & \multicolumn{2}{c}{Background}  & \multicolumn{2}{c}{Hand-Drill}  & \multicolumn{2}{c}{Backpack}    & \multicolumn{2}{c}{Fire-Extinguisher} & \multicolumn{2}{c}{Survivor}    & \multirow{2}{*}{mAcc} & \multirow{2}{*}{mIoU} \\ \cmidrule(lr){3-12}
		&                       & Acc            & IoU            & Acc            & IoU            & Acc            & IoU            & Acc               & IoU               & Acc            & IoU            &                       &                       \\ \midrule
		CCNet-4c                 & RGB                   & 99.62          & 98.95          & 74.62          & 55.93          & 86.09          & 78.82          & 65.15             & 49.24             & 57.51          & 52.77          & 76.60                 & 67.14                 \\
		DANet-4c                 & RGB                   & 99.74          & 99.02          & 82.80          & 68.28          & 83.28          & 78.49          & 58.16             & 48.05             & 54.73          & 50.46          & 75.74                 & 68.86                 \\ \midrule
		ACNet                    & RGBD                  & \textbf{99.83} & 99.25          & 53.59          & 51.46          & 85.56          & \textbf{83.19} & 84.88             & 59.95             & 69.10          & 65.19          & 78.67                 & 71.81                 \\
		SA-Gate                  & RGBD                  & 99.74          & 99.25          & 89.88          & \textbf{81.01} & 89.03          & 79.77          & 80.70             & 72.97             & 64.19          & 62.22          & 84.71                 & 79.05                 \\ \midrule
		MFNet                    & RGBT                  & 98.89          & 97.85          & 68.59          & 4.87           & 77.49          & 63.02          & 57.15             & 37.31             & 15.40          & 8.63           & 63.50                 & 50.34                 \\
		RTFNet-50                & RGBT                  & 99.74          & 98.89          & 64.07          & 52.24          & 73.08          & 67.91          & 69.78             & 54.46             & 58.19          & 54.11          & 72.97                 & 65.52                 \\
		PSTNet                   & RGBT                  & -              & 98.85          & -              & 53.60          & -              & 69.20          & -                 & 70.12             & -              & 50.03          & -                     & 68.36                 \\
		ABMDRNet                 & RGBT                  & 99.78          & 99.00          & 81.24          & 61.52          & 71.41          & 67.91          & 76.49             & 66.22             & 66.40          & 62.02          & 79.06                 & 71.33                 \\
		EGFNet                   & RGBT                  & 99.48          & 99.26          & \textbf{97.99} & 64.67          & \textbf{94.17} & 83.05          & \textbf{95.17}    & 71.29             & 83.30          & {74.30} & \textbf{94.02}        & 78.51                 \\
		MDRNet+                  & RGBT                  & 99.30          & 99.07          & 90.17          & 63.04          & 93.53          & 76.27          & 86.63             & 63.47             & \textbf{85.56} & 71.26          & 91.04                 & 74.62                 \\ \midrule
		Ours                     & RGBT                  & 99.70          & \textbf{99.36} & 93.45          & 73.14          & 87.48          & 81.99          & 92.74             & \textbf{77.66}    & 83.91          &  \textbf{74.88}          & 91.46                 & \textbf{81.41}        \\ \bottomrule
	\end{tabular}
\end{table*}

To further analyze the effectiveness of KL prior regularization, we design KL prior regularization with different prior information based on Eqn.(\ref{eq:prior}) for comparison, as shown in Table \ref{tbl:ablation-prior}. From the results, we can find that the segmentation performance under mIoU metric is significantly enhanced when considering both illumination and categories prior information. However, the segmentation performance is slightly worse when only considering prior of two illumination conditions (i.e., day and night,). The main reason is that the KL prior regularization of illumination is too coarse to introduce correct information for VFFM. The segmentation performance achieves about 0.1\% improvement when only considering category prior information, i.e., the number of Gaussian components is set to 9 in Eqn.(\ref{eq:prior}). Obviously, KL prior regularization based on Gaussian mixture distribution in Eqn.(\ref{eq:prior}) should have more Gaussian components, otherwise, it will not bring in significant improvement on segmentation performance. In the future work, we will try to further improve the modeling of KL prior regularization.

\subsection{Comparison with State-of-the-Arts}

\subsubsection{Comparison on MFNet Dataset}
We compare the proposed VPFNet with 13 state-of-the-art (SOTA) methods, including 2 deep learning RGB-based semantic segmentation methods (DANet~\cite{danet}, CCNet~\cite{huang2019ccnet}), 3 RGB-D semantic segmentation methods  (FuseNet~\cite{fusenet}, Depth-aware CNN~\cite{wang2018depth}, SAGate~\cite{chen2020bi}), and 8 RGB-T semantic segmentation methods (MFNet~\cite{MFNet}, RTFNet~\cite{RTFNet}, PSTNet~\cite{pstnet}, FuseSeg~\cite{fuseseg}, FEANet~\cite{FEANet}, ABMDRNet~\cite{ABMDRNet}, EGFNet~\cite{EGFNet}, MDRNet+~\cite{MDRNet}). 
The comparison results are shown in Table~\ref{tbl:MFNet-test} and Fig.\ref{fig:results}. From the results, we can find that although the proposed VPFNet does not show the best performance under most categories, it outperforms other state-of-the-art methods under mIoU metric when considering all categories, i.e., VPFNet has more balanced performance for different categories. Since the number of samples are not the same for different categories in the dataset, these results prove that VPFNet can simultaneously learn effective representations of all categories even if there exists class-imbalance \cite{imbalancereview} in the training data.  

Moreover, we summarizes the evaluation results from daytime and nighttime images respectively in Table~\ref{tbl:MFNet-daynight}. From the results, we can find that our proposed VPFNet still keep the state-of-the-art performance under mIoU metric. Additionally, the thermal images are more important at nighttime despite that it contains serious noise. However, compared with other methods, VPFNet performs far more better at nighttime, which proves that VPFNet is more robust to modality noise \cite{peng2022balanced} of thermal images. 

Besides, we discuss RGB-T semantic segmentation when losing an unknown modality. This scene occurs in realworld when one of RGB and thermal cameras is broken. To simulate this scene, we set the lost modality to zero and compare VPFNet with other methods, as shown in Table~\ref{tbl:MFNet-rgninf}. 
From the results, we can find that most segmentation methods based on the unique fusion feature assumption usually work badly without RGB images or thermal images, epspecially the FEANet~\cite{FEANet}. Although the proposed VPFNet does not show the best performance under a certain modality, it outperforms other methods when considering the worst performance under unknown lost modality. Thus, VPFNet has more balanced performance for different modalities, which prove that VPFNet is more robust to modality bias\cite{chen2020bi} and has powerful representation ability.

\subsubsection{Comparison on PST900 Dataset:}
We compare the proposed VPFNet with other state-of-the-art methods on the PST900 dataset~\cite{pstnet}. Different from the MFNet dataset, all images in PST900 dataset are taken under poor light condition. Thus, we only introduce category prior information for the proposed VPFNet in training process. The comparison results are summarized in Table~\ref{tbl:PST900}. From the results, we can find that VPFNet outperforms other methods under mIoU metric when considering all categories. Besides, VPFNet has more balanced performance for different categories compared with other methods. These results are similar to that on the MFNet dataset, which prove high generalization ability of VPFNet.

\section{Conclusion}
In this paper, we propose a novel VPFNet for RGB-T semantic segmentation. The proposed network regards fusion features as random variables, generates random samples of fusion feature via a novel VFFM based on variation attention, and obtains the robust segmentation by averaging all segmentation results of different fusion feature samples. Experiments on two public datasets prove that the proposed VPFNet has the state-of-the-art segmentation performance. Moreover, extensive experimental results show that VPFNet has more balanced segmentation performance for different categories and modalities, and it is more robust to class-imbalance, modality noise, and modality bias.


\ifCLASSOPTIONcaptionsoff
  \newpage
\fi



%
%
%

\bibliographystyle{IEEEtran}
\bibliography{egbib}

\begin{thebibliography}{10}
\providecommand{\url}[1]{#1}
\csname url@samestyle\endcsname
\providecommand{\newblock}{\relax}
\providecommand{\bibinfo}[2]{#2}
\providecommand{\BIBentrySTDinterwordspacing}{\spaceskip=0pt\relax}
\providecommand{\BIBentryALTinterwordstretchfactor}{4}
\providecommand{\BIBentryALTinterwordspacing}{\spaceskip=\fontdimen2\font plus
\BIBentryALTinterwordstretchfactor\fontdimen3\font minus
  \fontdimen4\font\relax}
\providecommand{\BIBforeignlanguage}[2]{{%
\expandafter\ifx\csname l@#1\endcsname\relax
\typeout{** WARNING: IEEEtran.bst: No hyphenation pattern has been}%
\typeout{** loaded for the language `#1'. Using the pattern for}%
\typeout{** the default language instead.}%
\else
\language=\csname l@#1\endcsname
\fi
#2}}
\providecommand{\BIBdecl}{\relax}
\BIBdecl

\bibitem{drive2}
F.~Nesti, G.~Rossolini, S.~Nair, A.~Biondi, and G.~Buttazzo, ``Evaluating the
  robustness of semantic segmentation for autonomous driving against real-world
  adversarial patch attacks,'' in \emph{proc. IEEE Winter Conf. on Applications
  of Comput. Vis.}, 2022, pp. 2280--2289.

\bibitem{drive1}
H.~Wang, Y.~Chen, Y.~Cai, L.~Chen, Y.~Li, M.~A. Sotelo, and Z.~Li, ``Sfnet-n:
  An improved sfnet algorithm for semantic segmentation of low-light autonomous
  driving road scenes,'' \emph{IEEE Trans. Intell. Transp. Syst.}, vol.~23,
  no.~11, pp. 21\,405--21\,417, 2022.

\bibitem{robot1}
I.~Alonso, L.~Riazuelo, and A.~C. Murillo, ``Mininet: An efficient semantic
  segmentation convnet for real-time robotic applications,'' \emph{IEEE Trans.
  Robotics}, vol.~36, no.~4, pp. 1340--1347, 2020.

\bibitem{robot2}
S.~Ainetter and F.~Fraundorfer, ``End-to-end trainable deep neural network for
  robotic grasp detection and semantic segmentation from rgb,'' in \emph{IEEE
  Int. Conf. on Robotics and Automation (ICRA)}.\hskip 1em plus 0.5em minus
  0.4em\relax IEEE, 2021, pp. 13\,452--13\,458.

\bibitem{MFNet}
Q.~Ha, K.~Watanabe, T.~Karasawa, Y.~Ushiku, and T.~Harada, ``Mfnet: Towards
  real-time semantic segmentation for autonomous vehicles with multi-spectral
  scenes,'' in \emph{IEEE In. Conf. on Intell. Robots and Systems
  (IROS)}.\hskip 1em plus 0.5em minus 0.4em\relax IEEE, 2017, pp. 5108--5115.

\bibitem{EGFNet}
W.~Zhou, S.~Dong, C.~Xu, and Y.~Qian, ``Edge-aware guidance fusion network for
  rgb--thermal scene parsing,'' in \emph{proc. AAAI Conf. Artif. Intell.},
  2022, pp. 3571--3579.

\bibitem{ABMDRNet}
Q.~Zhang, S.~Zhao, Y.~Luo, D.~Zhang, N.~Huang, and J.~Han, ``Abmdrnet:
  Adaptive-weighted bi-directional modality difference reduction network for
  rgb-t semantic segmentation,'' in \emph{proc. IEEE Conf. Comput. Vis. Pattern
  Recognit. (CVPR)}, 2021, pp. 2633--2642.

\bibitem{FEANet}
F.~Deng, H.~Feng, M.~Liang, H.~Wang, Y.~Yang, Y.~Gao, J.~Chen, J.~Hu, X.~Guo,
  and T.~L. Lam, ``Feanet: Feature-enhanced attention network for rgb-thermal
  real-time semantic segmentation,'' in \emph{IEEE In. Conf. on Intell. Robots
  and Systems (IROS)}.\hskip 1em plus 0.5em minus 0.4em\relax IEEE, 2021, pp.
  4467--4473.

\bibitem{doodlenet}
O.~Frigo, L.~Martin-Gaffe, and C.~Wacongne, ``Doodlenet: Double deeplab
  enhanced feature fusion for thermal-color semantic segmentation,'' in
  \emph{proc. IEEE Conf. Comput. Vis. Pattern Recognit. (CVPR)}, 2022, pp.
  3021--3029.

\bibitem{zhou2021gmnet}
W.~Zhou, J.~Liu, J.~Lei, L.~Yu, and J.-N. Hwang, ``Gmnet: graded-feature
  multilabel-learning network for rgb-thermal urban scene semantic
  segmentation,'' \emph{IEEE Trans. Image Processing}, vol.~30, pp. 7790--7802,
  2021.

\bibitem{MFFENet}
W.~Zhou, X.~Lin, J.~Lei, L.~Yu, and J.-N. Hwang, ``Mffenet: Multiscale feature
  fusion and enhancement network for rgb--thermal urban road scene parsing,''
  \emph{IEEE Trans. Multimedia}, vol.~24, pp. 2526--2538, 2021.

\bibitem{imbalancereview}
T.~Zhou, S.~Ruan, and S.~Canu, ``A review: Deep learning for medical image
  segmentation using multi-modality fusion,'' \emph{Array}, vol.~3, p. 100004,
  2019.

\bibitem{peng2022balanced}
X.~Peng, Y.~Wei, A.~Deng, D.~Wang, and D.~Hu, ``Balanced multimodal learning
  via on-the-fly gradient modulation,'' in \emph{proc. IEEE Conf. Comput. Vis.
  Pattern Recognit. (CVPR)}, 2022, pp. 8238--8247.

\bibitem{chen2020bi}
X.~Chen, K.-Y. Lin, J.~Wang, W.~Wu, C.~Qian, H.~Li, and G.~Zeng,
  ``Bi-directional cross-modality feature propagation with
  separation-and-aggregation gate for rgb-d semantic segmentation,'' in
  \emph{proc. Eur. Conf. Comput. Vis. (ECCV)}.\hskip 1em plus 0.5em minus
  0.4em\relax Springer, 2020, pp. 561--577.

\bibitem{VAE}
D.~P. Kingma and M.~Welling, ``Auto-encoding variational bayes,'' \emph{arXiv
  preprint arXiv:1312.6114}, 2013.

\bibitem{chen2021variational}
B.~Chen, Z.~Yan, K.~Li, P.~Li, B.~Wang, W.~Zuo, and L.~Zhang, ``Variational
  attention: Propagating domain-specific knowledge for multi-domain learning in
  crowd counting,'' in \emph{proc. IEEE Int. Conf. Comput. Vis. (ICCV)}, 2021,
  pp. 16\,065--16\,075.

\bibitem{breiman2001random}
L.~Breiman, ``Random forests,'' \emph{Machine learning}, vol.~45, pp. 5--32,
  2001.

\bibitem{lafferty2001conditional}
J.~Lafferty, A.~McCallum, and F.~C. Pereira, ``Conditional random fields:
  Probabilistic models for segmenting and labeling sequence data,'' in
  \emph{proc. IEEE int. conf. on Machine Learning (ICML)}, 2001, pp. 282--289.

\bibitem{dalal2005histograms}
N.~Dalal and B.~Triggs, ``Histograms of oriented gradients for human
  detection,'' in \emph{proc. IEEE Conf. Comput. Vis. Pattern Recognit.
  (CVPR)}, vol.~1.\hskip 1em plus 0.5em minus 0.4em\relax Ieee, 2005, pp.
  886--893.

\bibitem{lindeberg2012scale}
\BIBentryALTinterwordspacing
W.~Burger and M.~J. Burge, \emph{Scale-Invariant Feature Transform
  (SIFT)}.\hskip 1em plus 0.5em minus 0.4em\relax London: Springer London,
  2016, pp. 609--664. [Online]. Available:
  \url{https://doi.org/10.1007/978-1-4471-6684-9_25}
\BIBentrySTDinterwordspacing

\bibitem{FCN}
J.~Long, E.~Shelhamer, and T.~Darrell, ``Fully convolutional networks for
  semantic segmentation,'' in \emph{proc. IEEE Conf. Comput. Vis. Pattern
  Recognit. (CVPR)}, 2015, pp. 3431--3440.

\bibitem{unet}
O.~Ronneberger, P.~Fischer, and T.~Brox, ``U-net: Convolutional networks for
  biomedical image segmentation,'' in \emph{proc. Int. Conf. Medical Image
  Computing and Computer-Assisted Intervention (MICCAI)}.\hskip 1em plus 0.5em
  minus 0.4em\relax Springer, 2015, pp. 234--241.

\bibitem{ppm}
H.~Zhao, J.~Shi, X.~Qi, X.~Wang, and J.~Jia, ``Pyramid scene parsing network,''
  in \emph{proc. IEEE Conf. Comput. Vis. Pattern Recognit. (CVPR)}, 2017, pp.
  2881--2890.

\bibitem{deeplab}
L.-C. Chen, G.~Papandreou, I.~Kokkinos, K.~Murphy, and A.~L. Yuille, ``Deeplab:
  Semantic image segmentation with deep convolutional nets, atrous convolution,
  and fully connected crfs,'' \emph{IEEE Trans. Pattern Anal. Machine Intell.},
  vol.~40, no.~4, pp. 834--848, 2017.

\bibitem{danet}
J.~Fu, J.~Liu, H.~Tian, Y.~Li, Y.~Bao, Z.~Fang, and H.~Lu, ``Dual attention
  network for scene segmentation,'' in \emph{proc. IEEE Conf. Comput. Vis.
  Pattern Recognit. (CVPR)}, 2019, pp. 3146--3154.

\bibitem{huang2019ccnet}
Z.~Huang, X.~Wang, L.~Huang, C.~Huang, Y.~Wei, and W.~Liu, ``Ccnet: Criss-cross
  attention for semantic segmentation,'' in \emph{proc. IEEE Int. Conf. Comput.
  Vis. (ICCV)}, 2019, pp. 603--612.

\bibitem{RTFNet}
Y.~Sun, W.~Zuo, and M.~Liu, ``Rtfnet: Rgb-thermal fusion network for semantic
  segmentation of urban scenes,'' \emph{IEEE Robotics and Automation Letters},
  vol.~4, no.~3, pp. 2576--2583, 2019.

\bibitem{fuseseg}
Y.~Sun, W.~Zuo, P.~Yun, H.~Wang, and M.~Liu, ``Fuseseg: Semantic segmentation
  of urban scenes based on rgb and thermal data fusion,'' \emph{IEEE Trans.
  Automation Science and Engineering}, vol.~18, no.~3, pp. 1000--1011, 2020.

\bibitem{pstnet}
S.~S. Shivakumar, N.~Rodrigues, A.~Zhou, I.~D. Miller, V.~Kumar, and C.~J.
  Taylor, ``Pst900: Rgb-thermal calibration, dataset and segmentation
  network,'' in \emph{proc. IEEE int. conf. on robotics and automation
  (ICRA)}.\hskip 1em plus 0.5em minus 0.4em\relax IEEE, 2020, pp. 9441--9447.

\bibitem{resnet}
K.~He, X.~Zhang, S.~Ren, and J.~Sun, ``Deep residual learning for image
  recognition,'' in \emph{proc. IEEE Conf. Comput. Vis. Pattern Recognit.
  (CVPR)}, 2016, pp. 770--778.

\bibitem{densenet}
G.~Huang, Z.~Liu, L.~Van Der~Maaten, and K.~Q. Weinberger, ``Densely connected
  convolutional networks,'' in \emph{proc. IEEE Conf. Comput. Vis. Pattern
  Recognit. (CVPR)}, 2017, pp. 4700--4708.

\bibitem{li2022rgb}
G.~Li, Y.~Wang, Z.~Liu, X.~Zhang, and D.~Zeng, ``Rgb-t semantic segmentation
  with location, activation, and sharpening,'' \emph{IEEE Trans. Circuits and
  Syst. Video Technol.}, 2022.

\bibitem{guo2021robust}
Z.~Guo, X.~Li, Q.~Xu, and Z.~Sun, ``Robust semantic segmentation based on
  rgb-thermal in variable lighting scenes,'' \emph{Measurement}, vol. 186, p.
  110176, 2021.

\bibitem{liu2022gcnet}
J.~Liu, W.~Zhou, Y.~Cui, L.~Yu, and T.~Luo, ``Gcnet: Grid-like context-aware
  network for rgb-thermal semantic segmentation,'' \emph{Neurocomputing}, vol.
  506, pp. 60--67, 2022.

\bibitem{paszke2016enet}
A.~Paszke, A.~Chaurasia, S.~Kim, and E.~Culurciello, ``Enet: A deep neural
  network architecture for real-time semantic segmentation,'' \emph{arXiv
  preprint arXiv:1606.02147}, 2016.

\bibitem{betaVAE}
I.~Higgins, L.~Matthey, A.~Pal, C.~Burgess, X.~Glorot, M.~Botvinick,
  S.~Mohamed, and A.~Lerchner, ``beta-vae: Learning basic visual concepts with
  a constrained variational framework,'' in \emph{proc. Int. conf. on learning
  representations (ICLR)}, 2017.

\bibitem{deng2009imagenet}
J.~Deng, W.~Dong, R.~Socher, L.-J. Li, K.~Li, and L.~Fei-Fei, ``Imagenet: A
  large-scale hierarchical image database,'' in \emph{proc. IEEE Conf. Comput.
  Vis. Pattern Recognit. (CVPR)}.\hskip 1em plus 0.5em minus 0.4em\relax Ieee,
  2009, pp. 248--255.

\bibitem{wright2021ranger21}
L.~Wright and N.~Demeure, ``Ranger21: a synergistic deep learning optimizer,''
  \emph{arXiv preprint arXiv:2106.13731}, 2021.

\bibitem{fusenet}
C.~Hazirbas, L.~Ma, C.~Domokos, and D.~Cremers, ``Fusenet: Incorporating depth
  into semantic segmentation via fusion-based cnn architecture,'' in
  \emph{proc. Asian Conf. on Comput. Vis. (ACCV)}.\hskip 1em plus 0.5em minus
  0.4em\relax Springer, 2017, pp. 213--228.

\bibitem{wang2018depth}
W.~Wang and U.~Neumann, ``Depth-aware cnn for rgb-d segmentation,'' in
  \emph{proc. Eur. Conf. Comput. Vis. (ECCV)}, 2018, pp. 135--150.

\bibitem{MDRNet}
S.~Zhao, Y.~Liu, Q.~Jiao, Q.~Zhang, and J.~Han, ``Mitigating modality
  discrepancies for rgb-t semantic segmentation,'' \emph{IEEE Trans. Neural
  Networks Learn. Syst.}, 2023.

\end{thebibliography}

%

%
%
%




\end{document}